\documentclass{bmvc2k}
\usepackage{booktabs}
\usepackage{graphicx}
\usepackage{amssymb}

\title{Isolated Channel Vision Transformers:\\ From Single-Channel Pretraining to Multi-Channel Finetuning}

\addauthor{Wenyi Lian}{wenyi.lian@it.uu.se}{1}
\addauthor{Patrick Micke}{patrick.micke@igp.uu.se}{2}
\addauthor{Joakim Lindblad}{joakim.lindblad@it.uu.se}{1}
\addauthor{Nata\v{s}a Sladoje}{natasa.sladoje@it.uu.se}{1}

\addinstitution{
 Department of Information Technology\\
 Uppsala University\\
 Uppsala, Sweden
}
\addinstitution{
 Department of Immunology, Genetics and Pathology\\
 Uppsala University\\
 Uppsala, Sweden
}

\runninghead{Lian et al.}{Isolated Channel Vision Transformers (IC-ViT)}

\def\eg{\emph{e.g}\bmvaOneDot}

\begin{document}

\maketitle

\begin{abstract}

Vision Transformers (ViTs) have achieved remarkable success in standard RGB image analysis. However, applying ViTs to multi-channel imaging (MCI) data, \eg, for medical and remote sensing applications, remains a challenge. In particular, MCI data often consist of layers acquired from different modalities. Directly training ViTs on such data can obscure modality/channel-specific information and impair performance. 
In this paper, we propose \textit{Isolated Channel ViT (IC-ViT)}, a simple yet effective training framework for large-scale MCI data. By randomly sampling one channel per image per iteration, IC-ViT learns channel-specific representations without requiring multi-channel fusion during pretraining. These representations are later integrated during finetuning to capture cross-channel dependencies in downstream tasks.
Experiments on various benchmarks, including JUMP-CP and CHAMMI for cell microscopy, and So2Sat-LCZ42 for satellite imaging, show the proposed IC-ViT outperforms existing channel-adaptive approaches by 4–14\%. Moreover, its efficient training makes it a suitable candidate for large-scale pretraining of foundation models on heterogeneous data. Our code is available at \url{https://github.com/shermanlian/IC-ViT}.

\end{abstract}
    
\section{Introduction}
\label{sec:intro}
Self-supervised learning (SSL) of Vision Transformers (ViTs), as demonstrated by methods like DINO~\cite{caron2021emerging} and DINOv2~\cite{oquab2023dinov2}, has shown clear advantages~\citep{oquab2023dinov2,xu2024whole,awais2025foundation} in learning generalizable representations. The application of an SSL model typically consists of two stages: (1) pretraining on large-scale unlabeled data to learn general representations; and (2) fine-tuning on downstream datasets to address specific, task-oriented problems~\cite{ericsson2022self,zhou2024comprehensive}. However, a key limitation of ViT-based SSL models is that, in both pretraining and fine-tuning stages, they are typically designed for standard RGB images and cannot naturally handle inputs with more than three channels or heterogeneous channel configurations, as commonly encountered in multi-channel imaging (MCI)~\cite{cai2020review}.


A common strategy to address this challenge is to finetune large-scale pretrained ViTs on MCI data, typically by modifying the patch embedding layer to accommodate the varying number of input channels. However, directly transferring ViTs pretrained on RGB data to multi-modal MCI datasets often leads to mismatches in input representation~\cite{de2021efficient}, resulting in suboptimal exploitation of the available multi-channel information~\cite{zhou2024comprehensive}.
Recent works have thus shifted focus toward modifying the ViT architecture to support multi-channel inputs~\citep{savarese2019learning,bao2024channel,bourriez2024chada,pham2024enhancing,chen2023chammi}. These approaches, commonly referred to as \textit{channel-adaptive ViTs}, extract embeddings independently from each input channel and concatenate them into a single sequence to preserve channel-specific information. However, this expansion in sequence length substantially raises computational costs and significantly prolongs training time, posing a critical challenge in large-scale self-supervised learning~\cite{bao2024channel}.

\begin{figure}[t]
\setlength{\belowcaptionskip}{-0.1in}
    \begin{center}
    \includegraphics[width=1.\linewidth]{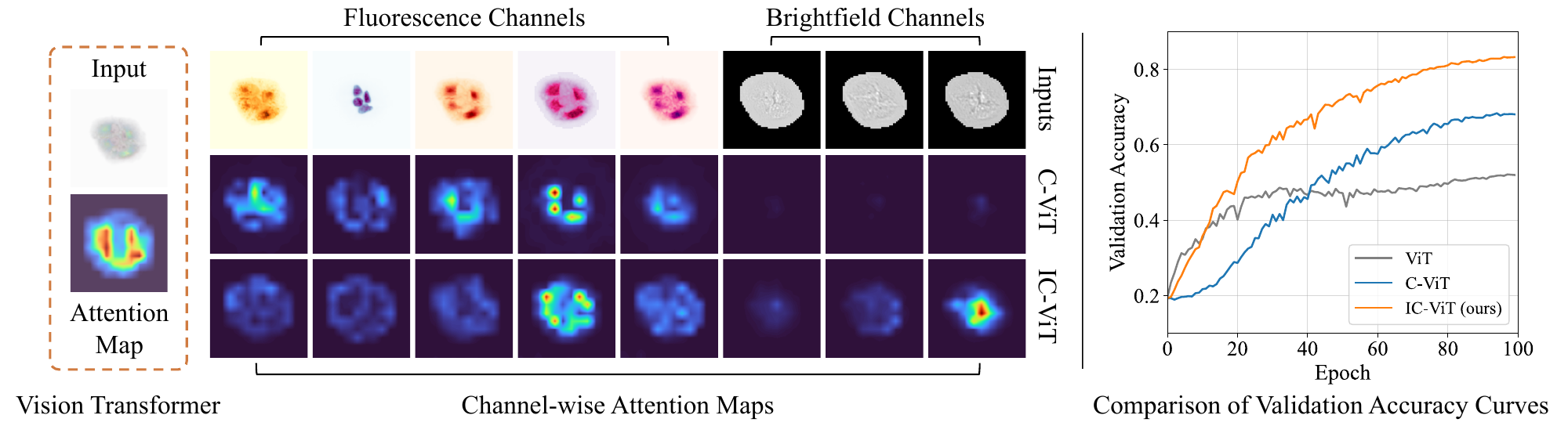}
    \caption{
        \textbf{Left:} Visualization of the attention maps generated from the standard Vision Transformer (ViT)~\cite{dosovitskiy2021an}, ChannelViT (C-ViT)~\cite{bao2024channel}, and the proposed \textbf{I}solated \textbf{C}hannel \textbf{ViT} (IC-ViT). Note that ViT compromises all channels into one patch token while ChannelViT and IC-ViT generate patch tokens for each channel individually. Our IC-ViT tends to extract information from both fluorescence and brightfield channels, while the latter is often ignored by ChannelViT. 
        \textbf{Right:} Performance comparison on the JUMP-CP validation dataset~\cite{chandrasekaran2024three}.
    }
    \label{fig:teaser}
    \end{center}
\end{figure}

In this paper, we present \textbf{Isolated Channel ViT (IC-ViT)}, a simple yet effective self-supervised learning method tailored for multi-channel imaging data. Specifically, IC-ViT is pretrained by learning representations from only a single channel per image at each iteration, rather than relying on all channels simultaneously. This isolated-channel strategy significantly reduces the computational cost of pretraining while enabling effective adaptation to multi-channel downstream tasks. While pretraining focuses on single-channel representations, fine-tuning on downstream tasks allows IC-ViT to integrate information across channels, leading to significantly better task performance.

As illustrated in \autoref{fig:teaser} (left), the attention maps generated by IC-ViT for an eight-channel JUMP-CP microscopy image indicate the model’s ability to capture informative features from both modalities. In addition, the comparison of training curves in the right panel of \autoref{fig:teaser} further confirms its effectiveness for downstream classification. Moreover, the scalability of IC-ViT to large-scale MCI datasets with diverse channel and modality compositions underscores its potential as a foundation model for MCI applications.



\section{Related work}
\label{sec:related_work}


\begin{figure}[t]
\setlength{\belowcaptionskip}{-0.1in}
    \centering
    \includegraphics[width=.62\linewidth]{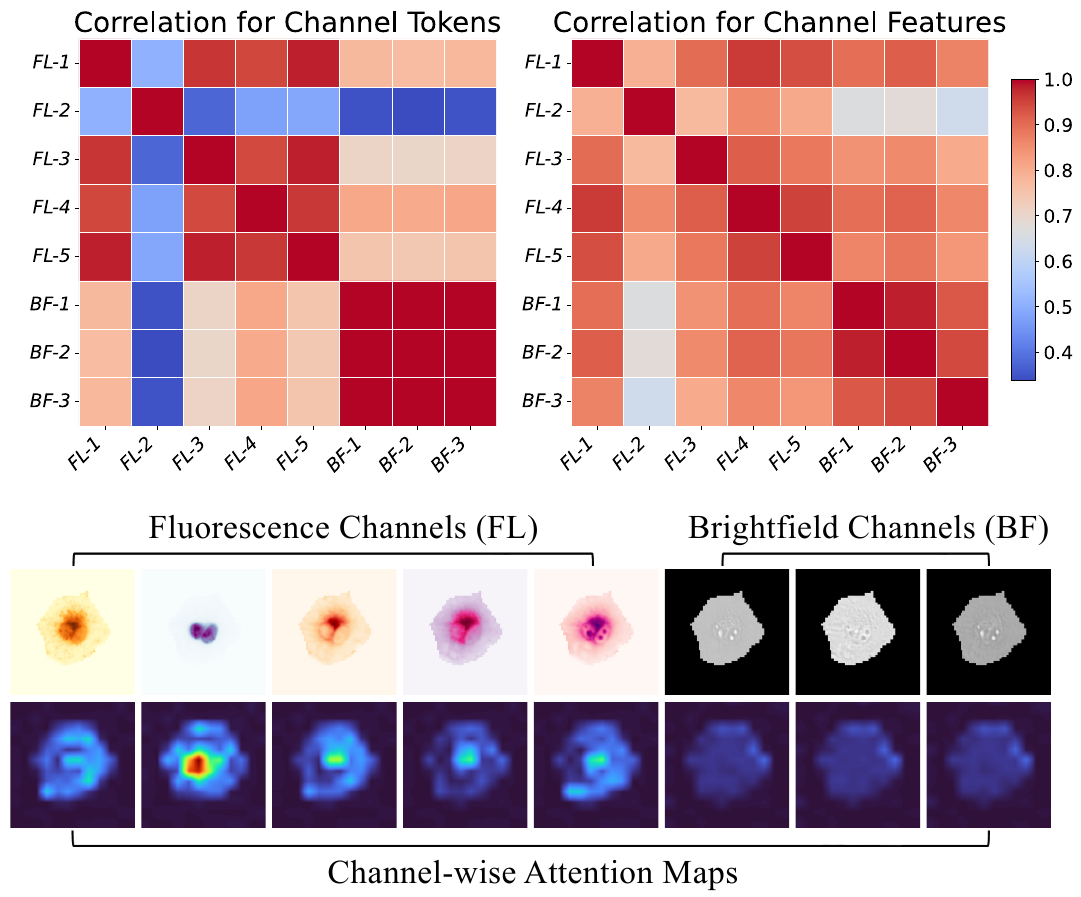}
    \caption{Inter-channel correlation analysis on the JUMP-CP dataset~\cite{chandrasekaran2024three}.
    \textbf{Top:} Pairwise correlations of channel tokens and features across eight microscopy channels. \textbf{Bottom:} Example of a multi-channel microscopy image and its channel-wise attention maps, obtained from a ViT pretrained with DINO on single-channel inputs.}
    \label{fig:corr_map}
\end{figure}



While ViTs~\cite{dosovitskiy2021an} have been highly successful, they have mainly been applied to RGB images~\cite{han2022survey,thisanke2023semantic}. However, multi-channel imaging is finding increased use in many fields, since it captures richer information beyond traditional RGB~\cite{misra2020multi,lee2019mcsip,liu2017deep,nie2019multi}.
This has led to the development of channel-adaptive models for multi-channel data~\cite{savarese2019learning, ha2016hypernetworks, chen2023chammi}. Recently, ClimaX~\cite{nguyen2023climax} is presented as a variant of the ViT that employs variable tokenization and aggregation methods to process heterogeneous weather and climate data. DepthwiseViT~\cite{chen2023chammi} proposes to process each input channel separately using a depthwise convolution layer. The extracted features are then aggregated by averaging, forming a new representation that is then passed into the ViT backbone. ChannelViT~\cite{bao2024channel} and ChAda-ViT\cite{bourriez2024chada} both generate patch tokens independently for each channel and incorporate learnable channel embeddings to explicitly differentiate between channels. Additionally, ChannelViT introduces Hierarchical Channel Sampling (HCS) during training to improve training efficiency. Building upon ChannelViT, DiChaViT~\cite{pham2024enhancing} further enhances feature diversity by adopting a Diverse Channel Sampling (DCS) strategy. DCS selectively samples less similar channels, reducing redundancy and mitigating over-similarity in channel representations.
However, existing channel-adaptive methods suffer from high computational costs due to channel-to-sequence expansion.


\section{Method}
\label{sec:method}


Multi-channel images often consist of layers acquired through different imaging protocols or modalities. A random example of an eight-channel image from the JUMP-CP dataset, including both fluorescence and brightfield microscopy channels, is shown at the bottom of \autoref{fig:corr_map}. These layers typically contain distinct yet complementary information that should be jointly leveraged for downstream analysis. In this section, we start from the ViT architecture and study \textit{why} channel-wise patchifying matters for multi-channel images and explore \textit{how} to efficiently extract informative features across channels. 

\begin{figure}[t]
    \centering
    \includegraphics[width=0.65\linewidth]{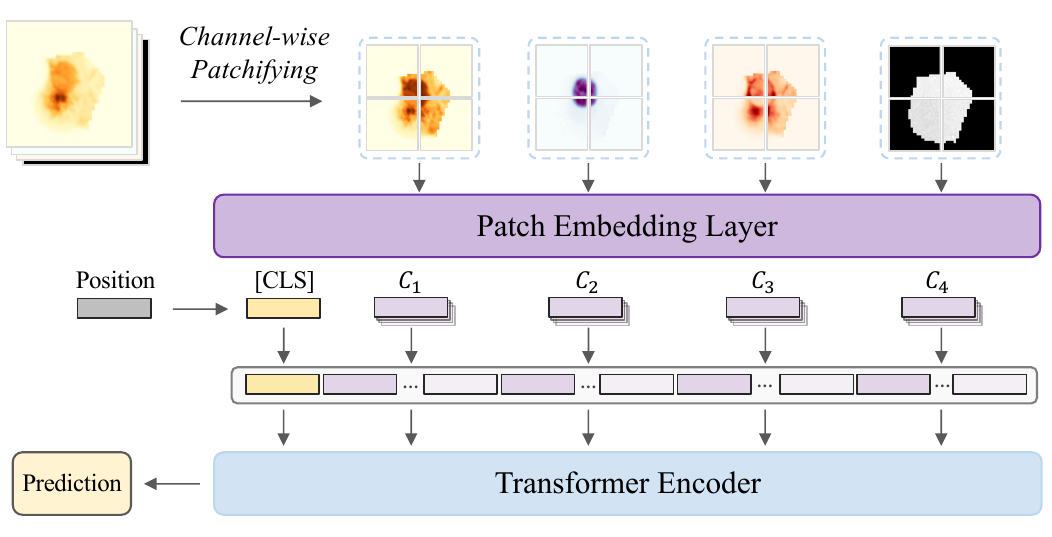}
    \caption{Illustration of channel-wise patchifying in ViT. Each channel is patchified individually, and the resulting patches are embedded into a sequence of vectors, with position and `\!\texttt{[CLS]}\!' embeddings added, to form the input to the transformer. $C_i$ denotes the embeddings of the $i$th channel.}
    \label{fig:vit_patchify}
\end{figure}

\subsection{Channel-wise Patchifying ViT}
\label{subsec:channel_patchify}

\paragraph{Preliminary}
Image patchifying is an important step in all vision transformers which splits the 2D image $\textbf{x} \in \mathbb{R}^{H \times W \times C}$ into $N$ patches $\textbf{x}_n \in \mathbb{R}^{P^2 \times C}$, where $P$ is the patch size and $N=HW/P^2$. ViT then maps each image patch into an embedding using a learnable linear projection $\mathbf{W}\in\mathbb{R}^{P^2C \times D}$ and concatenates all patches as a vector sequence $\mathbf{z}\in\mathbb{R}^{N \times D}$, where $D$ is the embedding dimension. In addition, a position embedding is added to retain positional information and a learnable classification token (`\!\texttt{[CLS]}\!') embedding is inserted at the start, to serve for final prediction. 

\paragraph{Correlation between channels}
 Before introducing the channel-wise patchifying ViT, we examine inter-channel correlations in MCI data. Specifically, we pretrain a single ViT with DINO~\cite{caron2021emerging} on single-channel inputs from the JUMP-CP microscopy dataset~\cite{chandrasekaran2024three} for self-supervised representation learning. The results are illustrated at the top of \autoref{fig:corr_map}, showing pairwise correlations of patch tokens and learned channel features across the eight channels in JUMP-CP.
 
We observe that tokens from fluorescence (FL) and brightfield (BF) channels exhibit lower correlations than tokens within the same modality, indicating that the two modalities provide complementary information. The high correlation observed among brightfield channels is largely due to their nearly identical visual appearance, which naïve token fusion cannot effectively differentiate. By isolating channels during pretraining, it is possible to extract subtle yet discriminative features from visually redundant channels, leading to more informative representations for downstream tasks. In contrast, projecting all channels into a single embedding is akin to ``early fusion'' in multimodal networks~\cite{ramachandram2017deep}, an approach that has been shown to be ineffective in many multimodal learning tasks~\citep{joze2020mmtm,he2023co,lian2025let}. Compared to patch tokens, the learned channel features exhibit stronger cross-modality correlations, suggesting that the model effectively learns shared representations bridging FL and BF channels. The attention maps in \autoref{fig:corr_map} (bottom) further confirm that the model attends to both FL and BF channels. These observations show the effectiveness of isolated-channel training and motivate channel-wise patchifying as a natural solution for multi-channel imaging~\citep{bao2024channel,pham2024enhancing}.

\paragraph{Channel-wise patchifying}
The main idea of channel-wise patchifying is to split the image into multiple single-channel images and patchify these images individually. The same projection layer is used for all channels. After projection, all channel patch embeddings are concatenated as the input sequence to the transformer. An overview of this process is shown in \autoref{fig:vit_patchify}. The resulting sequence $\tilde{\mathbf{z}}\in\mathbb{R}^{NC \times D}$ captures information from all channels and is $C$ times longer than the sequence in the original ViT. Moreover, a learnable channel embedding is often added to indicate the spatial index of each channel, similar to the position embedding in ViT~\cite{bao2024channel}. The transformer encoder then leverages self-attention to capture dependencies across patches and channels, yielding more robust feature representations.

\begin{figure}
    \centering
    \includegraphics[width=0.59\linewidth]{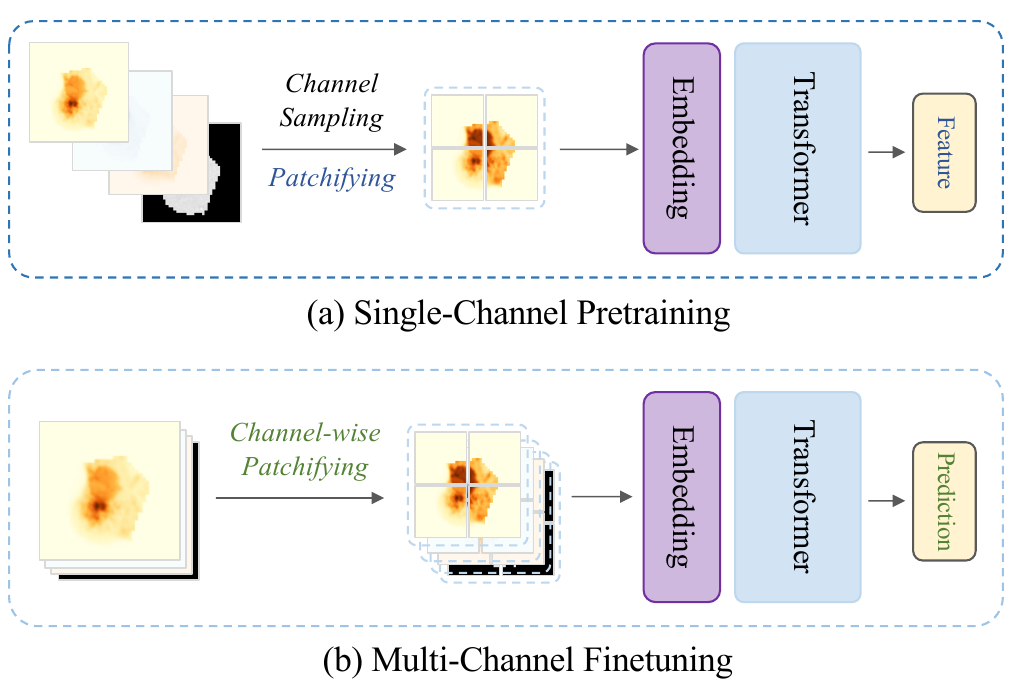}
    \caption{(a) Single-channel pretraining combined. (b) Multi-channel finetuning framework. IC-ViT samples single channel images for pretraining and uses all channels for prediction.}
    \label{fig:s2m_scheme}
\end{figure}

\subsection{Efficient Multi-Channel Pretraining}\label{subsec:MCI_pretraining}

We have demonstrated the importance of channel-wise patchifying in Sec.~\ref{subsec:channel_patchify} and illustrated the effectiveness of feature extraction with DINO pretraining in \autoref{fig:corr_map}. However, as noted earlier, channel-wise patchifying extends the sequence length by $C$ times, thereby leading to a quadratic growth in computational cost with respect to the number of channels due to the self-attention mechanism. This significantly limits the scalability of pretraining, particularly for self-supervised methods such as DINO~\cite{caron2021emerging} and DINOv2~\cite{oquab2023dinov2}, which process multiple augmented views per iteration and require dual networks (teacher and student) throughout training.

\paragraph{Single-channel pretraining} 
Training ViTs on individual channels, as described in Sec.~\ref{subsec:channel_patchify}, has empirically shown the ability to extract informative channel representations.
Since channel-wise patchifying introduces no channel-specific parameters, the number of channels can be adapted without modifying the network architecture (ViT naturally supports variable sequence lengths).
Then, the ViT pretrained on the individual channels can be finetuned on multi-channel images (MCI) for a downstream task. We refer to this approach as \textbf{I}solated \textbf{C}hannel \textbf{ViT} (IC-ViT). It is highly efficient for MCI pretraining and facilitates transfer to downstream tasks. An overview of the framework is presented in \autoref{fig:s2m_scheme}. In the pretraining stage, one channel is randomly sampled from each image and patchified as in a standard ViT for single-channel (grayscale) inputs. DINO~\cite{caron2021emerging} is employed to learn self-supervised representations from individual channels.  
During fine-tuning, channel-wise patchifying is applied to all channels simultaneously to leverage global cross-channel information.

\begin{figure}
\setlength{\belowcaptionskip}{-0.2in}
    \centering
    \includegraphics[width=0.6\linewidth]{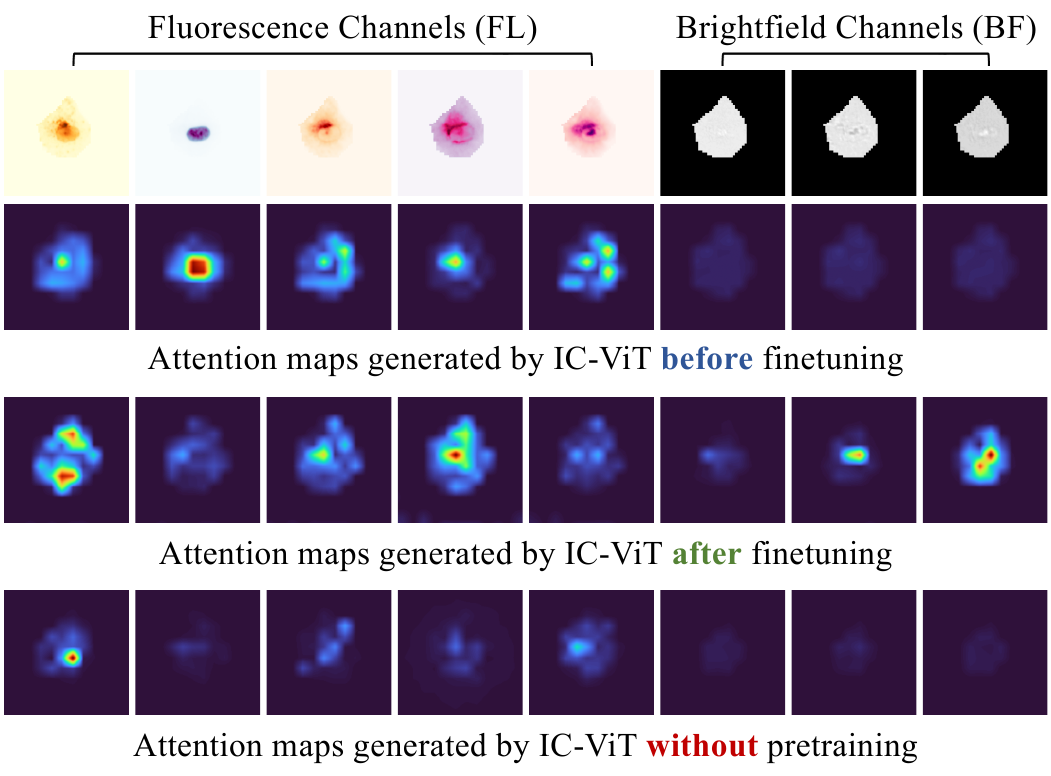}
    \caption{Attention maps on an eight-channel microscopy image from JUMP-CP dataset~\cite{chandrasekaran2024three}, generated by IC-ViT under different training settings: pretraining only (top), after supervised fine-tuning (middle), and trained directly with labels without pretraining (bottom).}
    \label{fig:attn-map-method}
\end{figure}

\autoref{fig:attn-map-method} illustrates the attention maps of IC-ViT under different training settings. Across all cases, the model attends more strongly to fluorescence channels, which typically contain clearer and more discriminative information than brightfield channels. Pretraining enables the model to also capture structural information from brightfield channels, which becomes better integrated with fluorescence features after finetuning. In contrast, models trained directly from scratch fail to adequately attend to brightfield channels, resulting in inferior performance compared to the finetuned model. Further details are provided in Sec.~\ref{sec:experiments}.

\paragraph{MCI foundation models}
With its efficient pretraining and finetuning scheme, the proposed IC-ViT is scalable to large MCI datasets with diverse channels and modalities. 
A key challenge in training models on multimodal multi-channel datasets is the difficulty of unifying heterogeneous image channels within the same training batch, which is required by most deep learning frameworks. Existing solutions typically adopt either (1) padding all images along the channel dimension, analogous to sentence padding in language models~\cite{sutskever2014sequence}, or (2) performing separate forward passes for different channel batches followed by gradient aggregation for parameter updates~\cite{pham2024enhancing}. 
However, both approaches are rigid and constrained by variability in channel numbers. 
In contrast, IC-ViT addresses this challenge by sampling a single channel from each image at a time, thereby ensuring channel uniformity within the batch while maintaining training efficiency.

\section{Experiments}
\label{sec:experiments}

In this section, we evaluate the proposed IC-ViT method on various tasks and MCI datasets.

\subsection{Datasets and Implementation}
\label{subsec:dataset}

We evaluate IC-ViT on three publicly available multi-channel benchmarks from the medical and remote sensing domains: 

1) \textit{JUMP-CP}~\cite{chandrasekaran2024three}, a widely used benchmark for cell microscopy imaging, with each image captured across eight channels (five fluorescence and three brightfield). Following previous works~\cite{bao2024channel,pham2024enhancing}, we use the compound perturbation plate `\texttt{BR00116991}', which comprises 127K training images, 45K validation images, and 45K test images. 

2) \textit{CHAMMI}~\cite{chen2023chammi}, a microscopy benchmark for channel-invariant models which combines three multi-channel imaging datasets with varying channel counts: WTC-11 hiPSC (WTC-11, 3 channels, 65K images)~\cite{viana2023integrated}, Human Protein Atlas (HPA, 4 channels, 67K images)~\cite{thul2017subcellular}, and Cell Painting (CP, 5 channels, 88K images)~\cite{way2022morphology}. 

3) \textit{So2Sat-LCZ42}~\cite{Zhu2020So2Sat}, a remote sensing dataset consisting of Sentinel-1 (8 radar channels) and Sentinel-2 (10 multispectral channels) imagery, with 352K training images, 24K validation images, and 24K test images. 

We apply the proposed single-channel sampling strategy for pretraining in all tasks. During both pretraining and fine-tuning, only the training data are used. For fine-tuning, we follow the same experimental setup as ChannelViT~\cite{bao2024channel} and DiChaViT~\cite{pham2024enhancing} to ensure fair comparison. For example, we use cross-entropy loss for JUMP-CP and So2Sat, and proxy-based loss~\cite{teh2020proxynca} for CHAMMI. For all downstream tasks, we employ the AdamW optimizer with a learning rate schedule that starts with a warm-up phase for the first 10 epochs.


For both JUMP-CP and So2Sat, the epoch achieving the highest validation accuracy is chosen, and the test set performance of that checkpoint is reported as the final top-1 accuracy. For CHAMMI, we follow the official CHAMMI evaluation protocol~\cite{chen2023chammi}, using the final checkpoint and reporting macro-F1 scores. Please refer to the Supplementary Materials for additional details on datasets, training settings, and results.


\subsection{Results}
\begin{table}[t]
\setlength{\belowcaptionskip}{-0.1in}
\begin{center}
\resizebox{.99\linewidth}{!}{
  \begin{tabular}{l @{\qquad} c c c @{\qquad} c c @{\qquad} c c}
    \toprule
    & \multicolumn{3}{c}{CHAMMI~\cite{chen2023chammi}\quad\null} 
    & \multicolumn{2}{c}{JUMP-CP~\cite{chandrasekaran2024three}\qquad\null} 
    & \multicolumn{2}{c}{\!So2Sat~\cite{Zhu2020So2Sat}} \\
    \cmidrule(r{2em}){2-4} \cmidrule(r{2em}){5-6} \cmidrule(r){7-8}
    \textbf{Model} 
    & {WTC} & {HPA}& {Avg F1-score} 
    & {Full} & {Partial} 
    & {Full} & {Partial} \\
    \midrule

    DepthwiseViT~\cite{chen2023chammi}& 50.35 &\textbf{71.52}&60.94&49.86&44.98&60.41&43.41\\
    TemplateMixingViT~\cite{savarese2019learning}& 49.51& 64.52 &57.02&52.48&43.85&55.86 &37.28\\
    HyperNetViT~\cite{ha2016hypernetworks}& 45.17& 63.90&54.54 &47.07 &42.43& 60.73& 41.88\\
    ChAda-ViT~\cite{bourriez2024chada} & 67.08 &60.67&63.88 &65.03& 42.15& 56.98& 12.38\\
    ChannelViT~\cite{bao2024channel} &67.66 &62.14 &64.90 &67.51 &56.49 &61.03& 46.16  \\
    DiChaViT~\cite{pham2024enhancing}& 73.36 & 63.35 & 68.36& 69.19 & 57.98 & 63.36 & 47.76 \\
    IC-ViT (Ours) & \textbf{74.45}&70.76&\textbf{72.54}&\textbf{83.43}&\textbf{66.01}&\textbf{67.10}&\textbf{48.21}\\
    \bottomrule
  \end{tabular}}
  \end{center}
  \caption{Comparison of method performance on test sets of three multi-channel image benchmarks: CHAMMI, JUMP-CP, and So2Sat. Best scores are shown in \textbf{bold}.}
  \label{tab:main_ret}
\end{table}

We compare the proposed IC-ViT to the following state-of-the-art channel-adaptive approaches: DepthwiseViT~\cite{chen2023chammi}, TemplateMixingViT~\cite{savarese2019learning}, HyperNetViT~\cite{ha2016hypernetworks}, ChAda-ViT~\cite{bourriez2024chada}, ChannelViT~\cite{bao2024channel}, and DiChaViT~\cite{pham2024enhancing}. For fairness, IC-ViT and all baseline methods are trained under the same experimental protocol, with identical task objectives, loss functions, and training splits. Importantly, IC-ViT employs the same number of parameters as ChannelViT, ensuring that performance differences are attributable to training strategies rather than model capacity.

The quantitative results are reported in \autoref{tab:main_ret}. For the JUMP-CP and So2Sat datasets, our evaluations cover two practical scenarios: (1) training and testing with all channels (denoted as ``Full''); and (2) training with all, but testing only with a partial set of channels (denoted as ``Partial''), simulating situations where only one modality is available at inference (the fluorescence channels for JUMP-CP and the Sentinel-1 channels for So2Sat). 
It is observed that the proposed IC-ViT consistently outperforms other approaches across most tasks and datasets. On the JUMP-CP dataset, IC-ViT achieves an improvement of 14–16\% over ChannelViT and DiChaViT in full-channel settings, and 8–10\% in partial-channel settings, demonstrating both its robustness to missing channels and its overall effectiveness. 
Similarly, on the CHAMMI benchmark, our method surpasses DiChaViT, the second-best method, by 4.18\% in overall F1-score, further confirming its strong capability for robust representation learning and generalization.

\section{Analysis and Discussion}
\label{sec:analysis}

\begin{figure}[t]
  \centering
  \begin{minipage}[t]{0.28\linewidth}
    \centering
    \includegraphics[width=1.\linewidth]{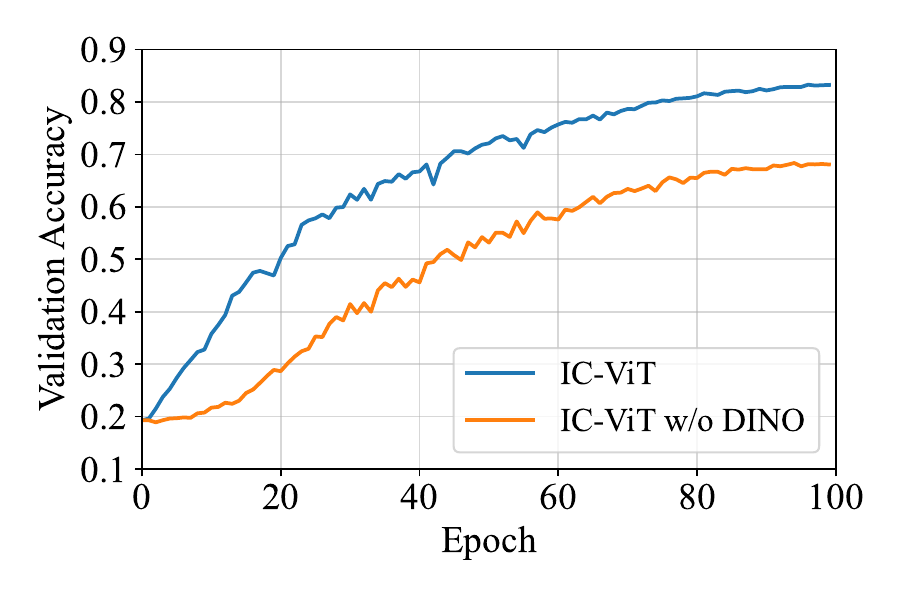}
  \end{minipage}
  \begin{minipage}[t]{0.28\linewidth}
    \centering
    \includegraphics[width=1.\linewidth]{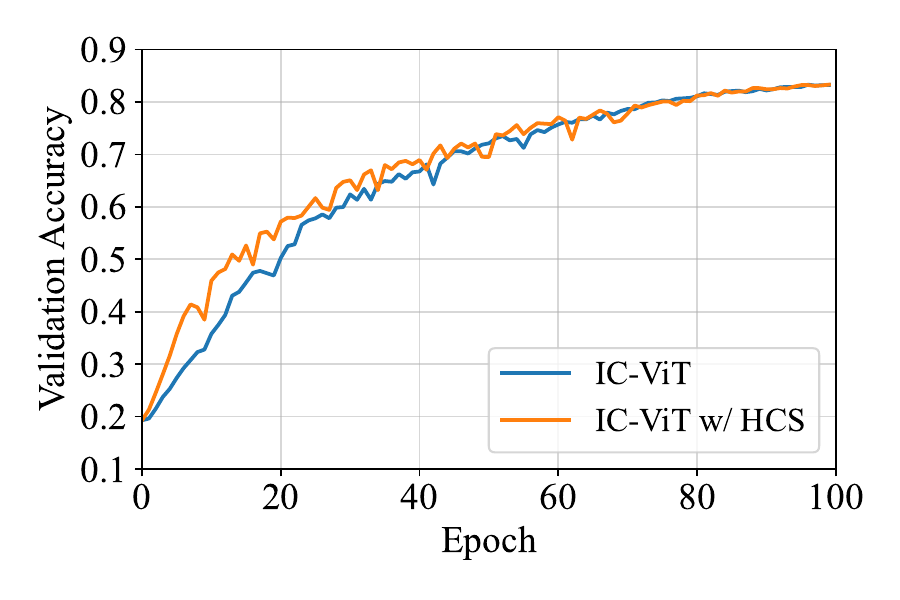}
  \end{minipage}
  \begin{minipage}[t]{0.35\linewidth}
  \vspace{-.9in}
    \begin{center}
    \resizebox{1.\linewidth}{!}{
  \begin{tabular}{@{\;}lcc@{\;}}  
    \toprule
    Model (S/16) & \#Channels & Time (hh:mm) \\
    \midrule
    ViT & 8  & 21:56    \\ 
    IC-ViT w/ HCS& $-$  & 86:28  \\ 
    IC-ViT & 8 & \!\!126:53 \\ 
    IC-ViT & 5  & 56:03  \\ 
    IC-ViT & 3  & 28:53\\ 
    IC-ViT (proposed) & 1 & 09:16\\ 
    \bottomrule
  \end{tabular}}
  \end{center}
\end{minipage}
\caption{(Left:) Validation accuracy for IC-ViT vs. IC-ViT without DINO pretraining. (Middle) Training curves of IC-ViT and its variant using HCS strategy~\cite{bao2024channel}. (Right:) Pretraining time (in hours:minutes) across different models and channel configurations on the JUMP-CP training set. Batch sizes are adjusted based on the available memory. The number of channels in ``IC-ViT w/ HCS'' is dynamically selected (denoted with ``$-$'').}
\label{fig:if-pre}
\end{figure}

\paragraph{Effectiveness of self-supervised learning}
We evaluate the impact of self-supervised pretraining in our method by comparing models trained with and without DINO on the JUMP-CP training set. As shown in \autoref{fig:if-pre} (left), the model pretrained with DINO significantly outperforms its supervised-only counterpart in downstream validation accuracy, highlighting the benefits of self-supervised representations.

  


\paragraph{Impact of channel sampling strategies on pretraining efficiency}
A major limitation of channel-adaptive approaches is the increased computational cost caused by expanding the channel dimension into the sequence length. We compare the pretraining times of ViT and IC-ViT variants employing different channel sampling strategies on the JUMP-CP training set over 100 epochs. All experiments are conducted on four 80GB A100 GPUs with consistent training configurations, except that the batch size varies with memory usage across channel settings.

As shown in \autoref{fig:if-pre} (right), pretraining IC-ViT on all channels (8 channels) without any sampling strategy takes approximately 127 hours (5 days and 7 hours). IC-ViT with Hierarchical Channel Sampling (HCS) instead samples a fixed number of channels at each iteration, thereby exposing the model to more channel information than our single-channel strategy. While incorporating HCS during DINO pretraining reduces the duration to 86 hours (3 days and 14 hours), Bao et al.~\cite{bao2024channel} reported that it can introduce instability into the self-distillation objective, which we also observe. More importantly, as shown in \autoref{fig:if-pre} (middle), although this variant achieves higher accuracy in the early stages of training, its advantage diminishes over time, resulting in comparable final performance to our method, which achieves similar results with substantially less training time (requiring only 9 hours). 



\paragraph{Effectiveness of channel-specific representation learning}

\begin{figure}[t]
  \centering
  \begin{minipage}[t]{0.28\linewidth}
    \centering
    \includegraphics[width=1.\linewidth]{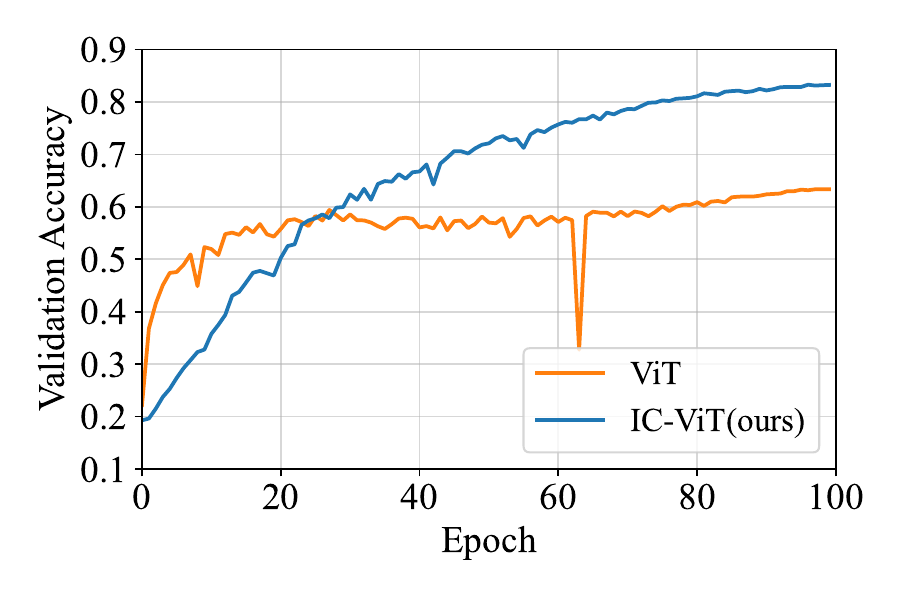}
  \end{minipage}
  \begin{minipage}[t]{0.28\linewidth}
    \centering
    \includegraphics[width=1.\linewidth]{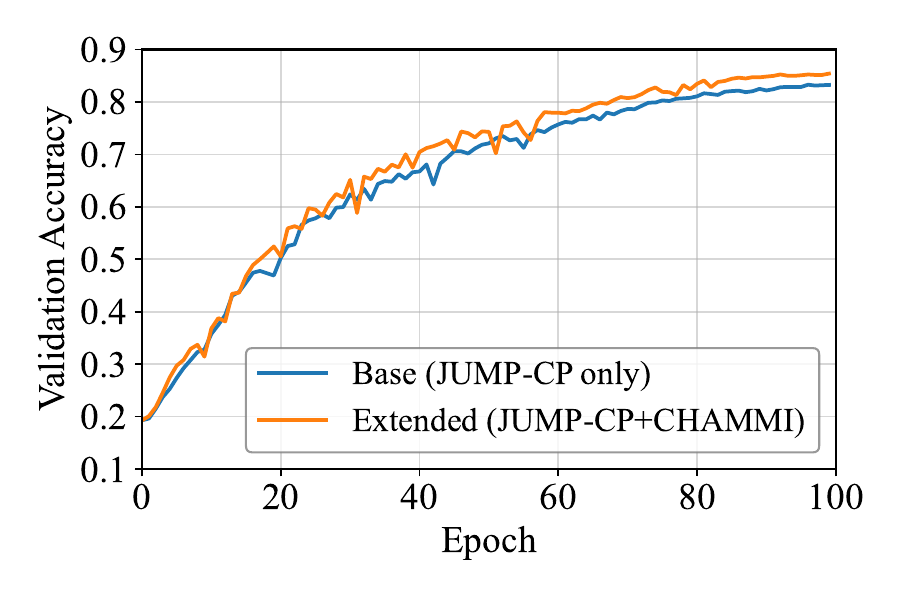}
  \end{minipage}
  \begin{minipage}[t]{0.35\linewidth}
  \vspace{-.9in}
    \centering
        \resizebox{\linewidth}{!}{
        \begin{tabular}{@{} l cc@{}}
            \toprule
             & Validation acc. & Test acc. \\
            \midrule
            {JUMP-CP}  & 83.25   & 83.41 \\
            {JUMP-CP+CHAMMI}  & \textbf{85.40} & \textbf{85.69} \\
            \bottomrule
        \end{tabular}
        }
\end{minipage}
\caption{(Left:) Performance curves of IC-ViT and ViT, both pretrained and finetuned on the JUMP-CP dataset. (Middle:) Multi-dataset pretraining evaluation. Validation accuracy curves for the baseline model (pretrained solely on JUMP-CP) and the extended model (pretrained on both JUMP-CP and CHAMMI). The extended pretraining generally results in higher accuracy throughout training. (Right:) Test accuracy at the epoch of highest validation accuracy. Best results in \textbf{bold}.}
\label{fig:others}
\end{figure}

The standard ViT~\cite{dosovitskiy2021an} treats all patches uniformly without differentiating among input channels, thereby neglecting the potentially independent semantic content carried by each channel/modality~\cite{bao2024channel}. 
We pretrain both ViT and IC-ViT using DINO on the JUMP-CP training set, and subsequently fine-tune them on the same dataset for the downstream classification task.  The corresponding training curves are shown in \autoref{fig:others} (left). While the standard ViT initially achieves higher accuracy, its performance stagnates after about 10 epochs. In contrast, IC-ViT continues to improve steadily, demonstrating the advantage of channel-specific representation learning.


\begin{table}[t]
    \begin{center}
    \resizebox{.63\linewidth}{!}{
    \begin{tabular}{@{}c@{\,}cccc@{}}
    \toprule
    \#channels \
      & \#comb.\
      & ViT \
      & ChannelViT \ 
      & IC-ViT (Ours) \ 
      \\
    \midrule
    8 &1&  56.87 &68.09 &83.41 \\
    7 &8& 49.35{\color{darkgray} \scriptsize$\pm$09.38} &61.02{\color{darkgray} \scriptsize $\pm$09.78}& 79.72{\color{darkgray}  \scriptsize $\pm$06.87} \\
    6 &28& 42.38{\color{darkgray} \scriptsize $\pm$10.64} & 53.45{\color{darkgray} \scriptsize $\pm$12.40}& 72.42{\color{darkgray} \scriptsize$\pm$10.11}\\
    5 &56& 35.78{\color{darkgray} \scriptsize$\pm$10.18} &45.50{\color{darkgray} \scriptsize$\pm$13.23} &  63.40{\color{darkgray} \scriptsize$\pm$12.31}\\
    4 &70&  29.84{\color{darkgray} \scriptsize$\pm$08.32} &37.37{\color{darkgray} \scriptsize$\pm$12.25}&  52.56{\color{darkgray} \scriptsize$\pm$13.10} \\
    3 &56& 24.94{\color{darkgray} \scriptsize$\pm$05.43} &29.68{\color{darkgray} \scriptsize$\pm$09.22}& 40.33{\color{darkgray} \scriptsize$\pm$11.71} \\
    2 &28& 21.54{\color{darkgray} \scriptsize$\pm$02.37} &23.77{\color{darkgray} \scriptsize$\pm$04.89}&  28.80{\color{darkgray} \scriptsize$\pm$07.45} \\
    1 &8&  19.92{\color{darkgray} \scriptsize$\pm$00.51}& 20.85{\color{darkgray} \scriptsize$\pm$01.64}&  21.84{\color{darkgray} \scriptsize$\pm$02.54} \\
    \bottomrule
    \end{tabular}
    }
    \end{center}
    \caption{Test accuracy on JUMP-CP~\cite{chandrasekaran2024three}. `\#comb.' denotes the number of $\binom{8}{k}$ combinations of channels. The results of ViT and ChannelViT are reported from their original paper~\cite{bao2024channel}.}
    \label{tab:channel-missing}
\end{table}



\paragraph{Robustness evaluation on channel subsets}

\autoref{tab:main_ret} shows that IC-ViT outperforms baseline methods even when an entire modality is absent at inference, indicating its potential for handling inputs with missing channels. We further extend this evaluation to scenarios with randomly missing channels by analyzing test accuracy on the JUMP-CP dataset when different subsets of channels are removed during inference. All possible channel combinations are considered, and \autoref{tab:channel-missing} reports the mean and standard deviation across these combinations for different methods. 

As illustrated in \autoref{tab:channel-missing}, IC-ViT consistently achieves higher average accuracy across different channel subsets, demonstrating strong robustness to partial or missing inputs. Moreover, the steady improvement in accuracy with more available channels indicates that our single-channel sampling strategy effectively leverages independent semantic information from each channel, enhancing the overall feature diversity and informativeness.


\paragraph{Pretraining with extended datasets}
As discussed in Sec.~\ref{subsec:MCI_pretraining}, IC-ViT has the potential to serve as a foundation model for MCI tasks by pretraining on extended datasets with a larger variety of multi-channel inputs. To provide a preliminary evaluation of this capability, we compare a baseline IC-ViT model pretrained solely on the JUMP-CP dataset with an extended version pretrained on both JUMP-CP and CHAMMI. Both models are subsequently fine-tuned on the JUMP-CP training set under the same experimental settings. As shown in \autoref{fig:others}  (middle), the model pretrained on the combined dataset generally leads to higher validation accuracy across most training epochs. As summarized in \autoref{fig:others} (right), pretraining with additional data results in over a 2\% improvement in both validation and test accuracy, further highlighting the model’s effectiveness for pretraining on heterogeneous multi-channel data.

\paragraph{Limitations and future work}
While IC-ViT demonstrates strong empirical performance across diverse MCI benchmarks, several limitations remain. First, the single-channel pretraining strategy inevitably neglects inter-channel correlations during representation learning, which may lead to partial information loss, especially in domains with low channel redundancy or high cross-channel dependencies. Second, our current uniform channel sampling treats all channels equally, which could be suboptimal in the presence of noisy or less informative channels. Future work could incorporate adaptive or learnable channel weighting mechanisms to better exploit heterogeneous data. Third, our experiments are limited to DINO-based self-supervised pretraining; extending the framework to other SSL paradigms such as MAE~\cite{he2022masked}, BYOL~\cite{grill2020bootstrap}, or SimMIM~\cite{xie2022simmim} would further validate its generality. Finally, the evaluation is restricted to microscopy and remote sensing datasets. Broader investigations on other multi-channel modalities, such as RGB-D imaging or medical MRI, will be essential to assess the scalability and universality of IC-ViT.

\section{Conclusion}
\label{sec:conclusion}

In this study, we tackle the challenge of learning from multi-channel imaging (MCI) data, where input channels often correspond to distinct modalities or measurements. We propose \textit{Isolated Channel Vision Transformer (IC-ViT)}, a simple yet effective framework that performs pretraining on single-channel inputs to learn channel-specific representations, while naturally supporting datasets with varying channel configurations. Upon fine-tuning, IC-ViT integrates complementary information across channels, leading to improved performance on downstream MCI tasks. Through extensive experiments on diverse datasets and tasks, we demonstrate that IC-ViT outperforms existing channel-adaptive approaches. Its efficiency and scalability make it a strong candidate for large-scale pretraining in MCI imaging.
\paragraph{Acknowledgement}
The computations and data handling were enabled by resources provided by Chalmers e-Commons at Chalmers.


\bibliography{main}

\end{document}